\def\year{2020}\relax
\documentclass[letterpaper]{article} 
\usepackage{aaai20}  
\usepackage{times}  
\usepackage{helvet} 
\usepackage{courier}  
\usepackage[hyphens]{url}  
\usepackage{graphicx} 
\usepackage{graphicx}  
\usepackage{amsmath}
\usepackage{algorithm}
\usepackage{algorithmic}
\usepackage{url}
\usepackage[subrefformat=parens]{subcaption}
\usepackage{comment}
\usepackage{amsfonts}

\nocopyright
\title{Joint Search of Data Augmentation Policies and Network Architectures}
\author{Taiga Kashima\textsuperscript{\rm 1}\thanks{This work was done during internship of Preferred Networks inc., Japan}, Yoshihiro Yamada\textsuperscript{\rm 2}, Shunta Saito\textsuperscript{\rm 2} \\
\textsuperscript{\rm 1} The University of Tokyo, \\
\textsuperscript{\rm 2} Preferred Networks inc., Japan}
\begin{document}

\maketitle

\begin{abstract}
The common pipeline of training deep neural networks consists of several building blocks such as data augmentation and network architecture selection.
AutoML is a research field that aims at automatically designing those parts, but most methods explore each part independently because it is more challenging to simultaneously search all the parts. 
In this paper, we propose a joint optimization method for data augmentation policies and network architectures to bring more automation to the design of training pipeline.
The core idea of our approach is to make the whole part differentiable.
The proposed method combines differentiable methods for augmentation policy search and network architecture search to jointly optimize them in the end-to-end manner.
The experimental results show our method achieves competitive or superior performance to the independently searched results.
\end{abstract}

\section{Introduction}
Neural network architectures that achieve high accuracy at specific tasks (e.g., image classification~\cite{vgg,resnet,densenet}) have been designed through many trials and errors by humans, which requires a high level of expertise and have been a burden to practitioners and researchers in the machine learning community.
Thus, AutoML has attracted the attention from many of them, which aims at automatically choosing a better design for each part in the machine learning pipeline, e.g., data augmentation~\cite{aa,fast_aa,pba,adv_aug}, network architecture~\cite{darts,nas_net}, loss functions~\cite{am_lfs}, or learning parameters~\cite{bayse,o_lr}.
In particular, several methods proposed in the field of Neural Architecture Search (NAS) started to achieve comparable accuracy to manually designed networks in a few tasks such as image classification~\cite{nas_net,proxyless_nas,pg_darts}.
However, the early methods of NAS typically have some practical problems mainly due to the large requirements for computing resources such as memories and GPUs~\cite{rl_nas}.
Then, recent studies tried to find better architecture with more efficient approaches~\cite{darts,enas,amoeba_net,ds_nas},
which lead the optimization goal of NAS to become varied.
\cite{fb_net} proposed a method to search network architectures that can achieve high accuracy under the limitation of computing resources used for inference,
and \cite{amc} proposed to consider memory and power efficiency at inference time by model compression.
Most of the methods mainly focused on the network architecture as their optimization target because it has large impact on the entire performance,
but as the optimization goal becomes diversified, the target of AutoML also becomes varied over different building blocks of the training,
e.g., data augmentation and learning parameters.
%

\label{intro}
\begin{figure}[!tbp]
  \centering
  \includegraphics[width=8cm,height=5cm]{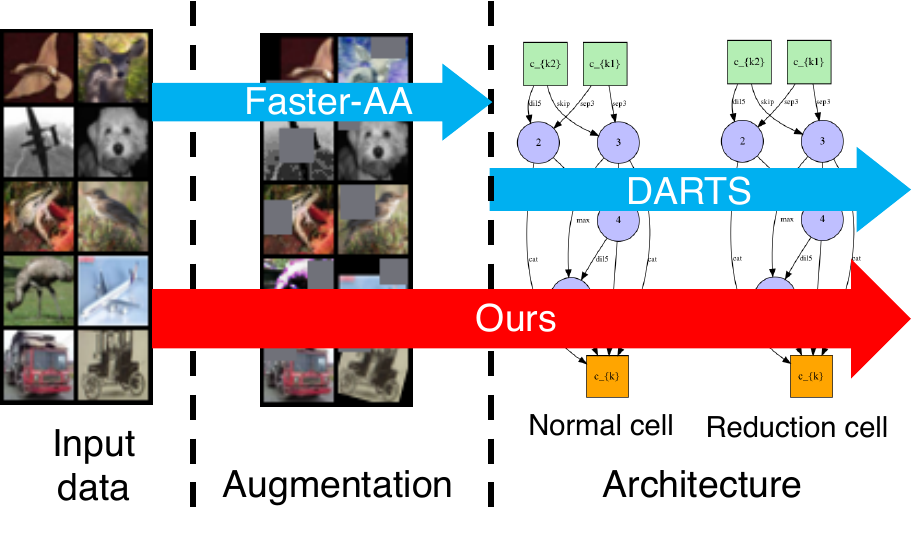}
  \caption{Comparison of search spaces. Our method jointly explores data augmentation policies and network architectures by combining differentiable methods for each part.}
  \label{search_space}
\end{figure}

In terms of search spaces, even a single part in the training pipeline has a large one.
For example, an efficient NAS method, ENAS~\cite{enas}, still has a large search space over $1.3 \times 10^{11}$ possible networks.
Auto Augment~\cite{aa} is a method to automatically choose the data augmentation policies during training, and the search space has roughly $2.9 \times 10^{32}$ possibilities.
It means that searching over all possibilities of the combination of these two parts will have about $3.8 \times 10^{43}$ possibilities, which brings significant difficulty to automatic exploration.
In addition, searching for network architectures and data augmentation policies can have different objectives.
The former explores mainly to minimize a loss function, while the latter also tries to increase the variety of training data.
These difficulties have prevented previous research from searching over those two parts jointly, so that most of the studies have focused on automatic exploration of one part at a time.
Although there are a few studies that tried to explore learning parameters and network architectures jointly~\cite{auto_has,toward_auto}, few research have attempted joint optimization of data augmentation policies and network architectures to the best of our knowledge.

In this paper, we propose a joint optimization method for both data augmentation policies and network architectures.
As stated above, these two parts can have large search spaces in total and different objectives, which implies that combining existing methods for each part straightforwardly would be intractable.
Additionally, during the progress of architecture search, networks in different phases of training can desire different data augmentation policies for better generalization ability.
Motivated by these intuitions, we propose an end-to-end differentiable approach to optimize both parts in the training pipeline simultaneously.
Fig.~\ref{search_space} shows the difference of our search space from previous successful methods for each part.
Specifically, we jointly optimize the differentiable approaches for augmentation policy search~\cite{faster_aa} and architecture search~\cite{darts}.
We firstly apply differentiable operations to the input data, and then use the transformation outputs as the inputs for the differentiable NAS method.
It enables to optimize the augmentation policies with the gradients come from the upstream NAS method because the entire pipeline is fully differentiable, so that we can train both parts simultaneously in the end-to-end manner.
We consider a combination of existing methods each of which is performed independently on either of augmentation policy search or architecture search as the baseline, and compare the performance with ours.
The experimental results show that our method achieves competitive or superior performance in common benchmarks of image classification.

\section{Preliminaries}
\label{preliminary}
As the differentiable methods for augmentation policy search and architecture search, we adopt Faster Auto Augmentation~\cite{faster_aa} and DARTS~\cite{darts}.
We briefly summarize them in this section.

\subsection{Differentiable Data Augmentation}
Data augmentation is a series of transformation applied on the input data.
Typically, we have to choose which operations should be applied with what magnitudes.
Several methods for automatic search of probability distribution on the selection of operators and their magnitudes have been proposed~\cite{aa,fast_aa,faster_aa}.
The Faster-AA considers that a policy consists of $L$ sub-policies each of which has $K$ consecutive operations.
Each operation out of $\#\mathcal{O}$ operations has a probability $p_O \in [0, 1]$ which represents how likely the operation is adopted in a sub-policy and a magnitude $\mu_O \in [0, 1]$ which controls the transformation.
Therefore, the search space is $(\#\mathcal{O} \times [0, 1] \times [0, 1])^{KL}$ in total, where $\mathcal{O}$ is a set of possible operations.
Applying an operation $O$ to an input data $X$ is formulated as:
\begin{eqnarray}
X \rightarrow  \left\{
\begin{array}{ll}
  O(X; \mu_O)& (\textrm{with the probability }p_O) \\
  X & (\textrm{with the probability of }1 - p_O), \label{diff_op}
\end{array}
\right.
\end{eqnarray}
and note that the Gumbel trick~\cite{gumbel} is used to make the probability differentiable.
A sub-policy is a series of $K$ operations, and during training, the output of $k$-th operation $X'$ from an input $X$ is calculated as a weighted sum over all possible $\#\mathcal{O}$ operations as follows:
\begin{eqnarray}
X' = \sum_{n=1}^{\#\mathcal{O}} [\sigma_\eta(\boldsymbol{z}_k)]_n O_k^{(n)}(X; \mu_k^{(n)}, p_k^{(n)}), \label{op1}\\
{\bf s.t.} \sum_{n=1}^{\#\mathcal{O}} [\sigma_\eta(\boldsymbol{z}_k)]_n = 1, \label{op2}
\end{eqnarray}
where $\sigma_\eta$ is a softmax function with
a temperature parameter $\eta > 0$,
and $\boldsymbol{z}_k \in \mathbb{R}^{\# \mathcal{O}}$ denotes the learnable parameter for the distribution of operator selection.
During inference, the $k$-th operation is sampled from the categorical distribution ${\rm Cat}(\sigma_\eta(\boldsymbol{z}_k))$, so that we obtain transformed data $X'$ by Eq.~(\ref{diff_op}).

\subsection{Differentiable NAS}
DARTS~\cite{darts} is a differentiable neural architecture search method which focuses on searching the inside structures of normal cells and reduction cells that are finally stacked up to build a deep network. 
Each cell is represented as a directed acyclic graph (DAG) consisting of $N$ nodes which represent intermediate features (e.g., feature maps).
A cell takes two input nodes and one output node, and
an edge $f^{(i, j)}$ between two nodes $i, j$ represents an operation such as convolution or pooling.
A node $j$ has the connections with all the previous nodes $i < j$ in topological ordering of the DAG, so that the search space of DARTS is roughly
\begin{eqnarray}
|\mathcal{F}|^2 \prod_{k=1}^{N-3}\frac{k(k+1)}{2},
\end{eqnarray}
where $\mathcal{F}$ is a set of candidate operations.
To make the search space continuous, DARTS relaxes the categorical choice of an operation to a softmax over possible operations:
\begin{eqnarray}
\bar{f}^{(i,j)} = \sum_{f \in \mathcal{F}} \frac{\exp(\alpha_f^{(i,j)})}{\sum_{f'} \exp(\alpha_{f'}^{(i,j)})}f(x).
\end{eqnarray}
After the training, a discrete architecture is determined by replacing each mixed operation $\bar{f}^{(i, j)}$ with the most likely operation $f^{(i, j)} = \mathop{\rm arg~max}\limits_{f \in \mathcal{F}} \alpha_f^{(i, j)}$.

DARTS performs a bilevel optimization on the architecture parameter $\alpha$ and the model weights $w$ using two sets of training and validation data.
\begin{eqnarray}
  \underset{\alpha}{\rm min}\ \mathcal{L}_{val} (w^*(\alpha), \alpha), \label{darts1} \\
  {\bf s.t.}\  w^*(\alpha) = \underset{w}{\rm arg~min}~\mathcal{L}_{train}(w, \alpha), \label{darts2}
\end{eqnarray}
where $\mathcal{L}_{val}$ and $\mathcal{L}_{train}$ are loss functions calculated with validation data and train data, respectively.
$w^*(\alpha)$ denotes the optimal model weights for an architecture $\alpha$.
Eq. (\ref{darts1}) has an inner optimization for $w^*({\alpha})$, so that evaluating the gradient of $\mathcal{L}_{val}$ w.r.t. $\alpha$ can be prohibitive.
Therefore, DARTS approximates $w^*(\alpha)$ by adapting $w$ only after a single training step on the training data as follows:
\begin{eqnarray}
  w^*(\alpha) = w - \xi\nabla_w \mathcal{L}_{train}(w, \alpha) \label{darts3}.
\end{eqnarray}
Then, during a single step in the iterative optimization procedure of DARTS, it solves Eq. (\ref{darts1}) and Eq. (\ref{darts2}) alternately.

\section{Method}
\label{method}
We sequentially combine Faster-AA and DARTS, then optimize both in the end-to-end manner.
We solve another bilevel optimization problem for our entire search space.
Specifically, augmentation policies and network architectures are both optimized by minimizing a loss function on the validation dataset, while the network weights are optimized using the training dataset.
This bilevel optimization is formulated as:
\begin{eqnarray}
\underset{\alpha,\boldsymbol{z}_k,p_O,\mu_O}{\rm min}\ \mathcal{L}_{val}(w^*(\alpha),\alpha,\boldsymbol{z}_k,p_O,\mu_O), \label{eq1}\\
{\bf s.t.}\ w^*(\alpha) = \underset{w}{\rm arg~min}~\mathcal{L}_{train}(w,\alpha,\boldsymbol{z}_k,p_O,\mu_O), \label{eq2}
\end{eqnarray} 
and we adopt the first-order approximation for the architecture gradient for Eq. (\ref{darts1}) same as in DARTS to speed-up the optimization, i.e., we set $\xi$ to $0$ in Eq. (\ref{darts3}).
Then, we iteratively solve Eq. (\ref{eq1}) and Eq. (\ref{eq2}).
As a loss function, we use the cross-entropy loss for both $\mathcal{L}_{train}$ and $\mathcal{L}_{val}$.

To solve Eq. (\ref{eq1}), we first apply a series of differentiable data augmentation operations $O^{(n)} (n=1,...,\#\mathcal{O})$, then give the transformed data to the network to solve Eq. (\ref{eq2}).
We outline this algorithm in Alg. \ref{alg1}.

\begin{algorithm} 
  \caption{Joint optimization}\label{alg1}
\begin{algorithmic}[1]
  \WHILE{\textit{not converged}}
  \STATE // solve equation (\ref{eq1})
  \STATE sample $X \sim D_{val}$ \\
  \STATE apply equation (\ref{op1}) to $X$
  \STATE calculate $L_{val}(w, \alpha, \boldsymbol{z}_k, p_O, \mu_O)$ with $X'$ \\
  \STATE calculate gradients of $L_{val}$ w.r.t. $\boldsymbol{z}_k, p_O, \mu_O, \alpha$ \\
  \STATE update Faster-AA parameters ($z_k, p_O, \mu_O$)
  \STATE update DARTS parameters $\alpha$ by gradient descent \\
  \STATE // solve equation (\ref{eq2})
  \STATE sample $X \sim D_{train}$ \\
  \STATE apply equation (\ref{op1}) to $X$ \\
  \STATE calculate $L_{train}(w, \alpha, \boldsymbol{z}_k, p_O, \mu_O)$ with $X'$ \\
  \STATE calculate gradients of $L_{train}$ w.r.t. $w$ \\
  \STATE update the network weights $w$ by gradient descent \\
  \ENDWHILE
\STATE Derive the final policy and architecture
\end{algorithmic}
\end{algorithm}

The total computational resources we require for the joint optimization additionally to what DARTS requires are relatively small.
Specifically, the additional space complexity is only $KL(\#\mathcal{O} \times 3)$, while the search space of Faster-AA is $(\#\mathcal{O} \times [0, 1] \times [0, 1])^{KL}$ which is large.
In addition to it, our entire system is end-to-end differentiable, and augmentation policy search and network architecture search are jointly performed to minimize the same loss function, so that the gradients for updating policy and architecture parameters are obtained via a single backpropagation.
This advantage of our end-to-end differentiable approach enables to conduct joint optimization for policy search and architecture search with few additional space and time complexity compared to the case if we apply Faster-AA and DARTS independently and combine the results.

The original Faster-AA uses a critic network to consider a classification loss and the WGAN-GP loss~\cite{wgan_gp} that encourage the distribution of transformed data to be as close to the original data distribution as possible.
For the same purpose, we can also exploit the network under searching with DARTS as a critic network to encourage transformed data to remain in the same classes before transformations.
However, DARTS only uses a cross-entropy as the loss for architecture search, while the critic network in Faster-AA also considers WGAN-GP.
In this paper, we adopt a single unified loss function for the both of policy search and architecture search and did not introduce any critic network for simplicity of the entire framework and computational efficiency.

\section{Experiments}
\label{experiments}
We compare our joint optimization model with the original DARTS and the baseline which combines the results of Faster-AA and DARTS that are optimized independently from each other.
In the baseline, the learned policy by Faster-AA is transferred to be used for the training of DARTS.
We conducted the comparison on three datasets, CIFAR-10, CIFAR-100, and SVHN.

\label{results}
\begin{figure*}[htbp]
\begin{minipage}{0.3\hsize}
  \centering
  \includegraphics[width=5cm,height=4cm]{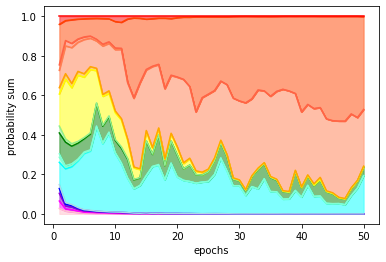}
  \subcaption{On CIFAR-10}
\end{minipage}
\begin{minipage}{0.3\hsize}
  \centering
  \includegraphics[width=5cm,height=4cm]{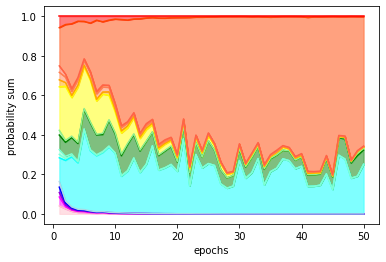}
  \subcaption{On CIFAR-100}
\end{minipage}
\begin{minipage}{0.3\hsize}
  \centering
  \includegraphics[width=6.5cm,height=4cm]{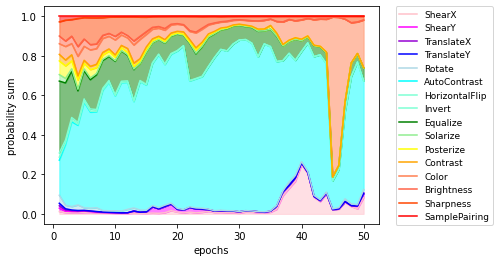}
  \subcaption{On SVHN}
  \label{svhn_policy}
\end{minipage}
  \caption{
  Probability distribution of the augmentation policy selection over time.
  }
  \label{res1}
\end{figure*}

We first carefully re-implemented Faster-AA\footnote{source code from: \url{https://github.com/moskomule/dda/tree/fasteraa/faster_autoaugment}} and DARTS\footnote{source code from: \url{https://github.com/quark0/darts}} based on their authors' implementation.
Then, we confirmed our implementation successfully reproduced the reported scores in those papers.
It should be noted that we adopt the same search space as the original paper for DARTS, while we exclude the cutout operation from the search space for the Faster-AA part in our framework as the authors do in their implementation.
According to a comment which has been left in their code, the cutout operation makes the optimization unstable.
Therefore, the target operations in the augmentation policy search are \textit{shear X, shear Y, translate X, translate Y, rotate, auto contrast, horizontal flip, invert, equalize, solarize, posterize, contrast, color, brightness, sharpness, and sample pairing}.
We set the number of sub-policies $L = 10$ and the number of operations in a sub-policy $K = 2$, which are the same in the original settings.
Additionally, as preprocessings in the baseline, random cropping with zero-padding and random horizontal flipping are always applied.
After those preprocessings, we apply transformations under the policy search with Faster-AA.
The cutout operation is always applied after the Faster-AA part in the baseline, although it is not included in the search space.

For the baseline, we first obtain a Faster-AA policy independently from DARTS by following the same experimental settings used in the original paper, i.e., the baseline uses WideResnet40-2~\cite{wideresnet} as the architecture to search policies for 20 epochs.
Next, the network architecture for the baseline is explored by DARTS in the same manner with the original paper.
Those policy and architecture obtained by the existing methods are combined together, then we train the model weights for 600 epochs.
As for the searching epochs of architectures, we conduct two different total epochs (25 and 50) both for the baseline and proposed framework.
Because we found that training of DARTS is unstable on CIFAR-100 and SVHN as mentioned in~\cite{darts_stab}.
\textit{We repeated experiments three times to see average scores with standard deviations.}

\subsection{Discussion}
\begin{table}[htbp]
  \begin{center}
  \caption{
  Comparison in classification accuracy.
  (1) ``DARTS" only searches architecture.
  (2) ``Baseline" separately searches policies and architectures.
  (3) ``Ours" is the proposed joint search method.
  Our method achieves competitive or superior results compared with the baseline.}
  \label{tab1}
  \begin{tabular}{c|c|c|c} \hline
    method & CIFAR-10 & CIFAR-100 & SVHN \\ \hline
    \multicolumn{4}{c}{searching epoch is 50} \\ \hline 
    DARTS & 97.33$\pm$0.12 & 76.21$\pm$3.76 & 97.85$\pm$0.08 \\
    Baseline & \textbf{97.55$\pm$0.30} & 77.96$\pm$3.12 & \textbf{98.02$\pm$0.03} \\
    Ours & 97.40$\pm$0.03 & \textbf{79.02$\pm$2.14} & 97.92$\pm$0.12  \\ \hline
    \multicolumn{4}{c}{searching epoch is 25} \\ \hline
    DARTS & 97.14$\pm$0.04 & 82.91$\pm$0.30 & 97.94$\pm$0.06 \\
    Baseline & 97.29$\pm$0.03 & \textbf{84.17$\pm$0.29} & \textbf{98.03$\pm$0.05} \\
    Ours & \textbf{97.46$\pm$0.09} & 83.81$\pm$0.49 & 97.82$\pm$0.08 \\ \hline
  \end{tabular}
  \end{center}
\end{table}


We present the experimental results in Table \ref{tab1}.
Our proposed method achieves competitive or superior results compared to the baseline in both searching epochs.
As stated above, the baseline uses WideResnet40-2 during policy search, and it can be different from the architecture found by DARTS for the final stage to learn the model weights for 600 epochs. 
Although the policies found with WideResnet40-2 show high accuracy at several results, which might be derived from the suitability of WideResnet40-2 for augmentation policy search.
If that is the case, the model selection largely affects on the policy search, so that the human expertise is still required.
On the other hand, our proposed method does not require humans to select the network architecture for policy search and achieves competitive or even superior results to Faster-AA.
We believe that the joint optimization approach has more potential to obtain high performance and reduces human expertise. 

Fig. \ref{res1} shows how the categorical distribution to choose augmentation policies changes over the time.
We found that the augmentation policies obtained with our method choose color enhancement operations such as \textit{color} or \textit{auto contrast} more often than geometric operations such as \textit{rotate} from the transition.
This trend of the resulting policies is also reported in other papers of automatic augmentation policy search~\cite{aa}.
In Figure \ref{svhn_policy}, the policy dramatically changes after 45 epochs, and we found that the network architecture under searching also largely changed at the same timing, which may imply that the optimal policy is different depending on the network architecture.

\section{Conclusion}
\label{conclusion}
In this paper, we proposed a method to jointly optimize data augmentation policy and network architecture.
The proposed method combines differentiable methods for policy search and architecture search to jointly optimize them in the end-to-end manner.
The experimental results showed that our method achieves competitive or superior performance to independently searched results in common benchmarks of image classification.
Our joint optimization approach may be able to include the other parts such as learning rates.
Hence, we will attempt to bring more automation to the design of training pipeline with this end-to-end differentiable approach for the future work.
\bibliographystyle{aaai}
\bibliography{pfn_ref}

\end{document}